%% file: main_iclr_arxiv.tex
\documentclass{article}
\usepackage{iclr2027_preprint,times}
\iclrfinalcopy
\usepackage{amsmath,amssymb,graphicx,booktabs,multirow,makecell}
\usepackage{microtype,float}
\usepackage{hyperref,url}

\newcommand{\method}{\textsc{ConMoE}}
\newcommand{\E}{\mathbb{E}}
\title{ConMoE: MoE Compression as\\Expert Selection and Functional Assignment}

\author{
\textbf{Yilun Yao,\hspace{0.4em} JiaMing Pan,\hspace{0.4em} Elsie Dai,\hspace{0.4em} Peizhuang Cong,\hspace{0.4em} Chao-Hsuan Liu,}\\
\textbf{Yaoming Li,\hspace{0.4em} Tong Yang}\thanks{
Corresponding author: Tong Yang. For questions about this work, please contact
\texttt{yilunyao@stu.pku.edu.cn} or
\texttt{yangtong@pku.edu.cn}.
}\\[0.6em]
\normalfont Peking University
}

\begin{document}
\maketitle
\input{Body/0.abstract}
\input{Body/1.intro}
\input{Body/5.related}
\input{Body/2.formulation}
\input{Body/3.method}
\input{Body/4.experiments}

\input{Body/6.conclusion}
\clearpage
\subsection*{AI use statement}
We use AI tools to assist with literataure research, methodological discussions, and manuscript polishing. All results have been manually reviewed by the authors.
\subsection*{Reproducibility statement}
Appendix~\ref{app:implementation} specifies the calibration and evaluation procedure, assignment settings, and baseline adaptations. Appendix~\ref{app:transfer} provides additional experimental results, assignment controls, and uncertainty estimates. Appendix~\ref{app:fingerprint} details fingerprint retrieval and its accuracy and construction cost relative to full-parameter retrieval.

\subsection*{Acknowledgments}
This work was supported by the Beijing Natural Science Foundation (Grant No. QY24044). The authors gratefully acknowledge this financial support, which helped facilitate the research activities and experimental investigations presented in this paper.

\bibliography{references}
\bibliographystyle{iclr2027_conference}
\clearpage
\appendix
\raggedbottom
\input{Body/appendix}
\end{document}

%% file: Body/0.abstract.tex
\begin{abstract}
Sparse Mixture-of-Experts (MoE) models scale model capacity through sparse activation, but storing their full parameters remains costly in GPU memory. Existing expert compression methods address this problem through pruning, merging, or substitution, typically coupling expert pool construction with the assignment of routed workloads. In this work, we formulate expert compression as two decisions---constructing an expert pool and assigning the original experts' workloads to that pool---and show that assignment can be optimized after the pool is fixed. Based on this formulation, we introduce \method, which constructs a compact pool using Anchor Coverage and optimizes assignment through a router-conditioned functional objective derived from a layer reconstruction bound, treating parameter proximity only as a proxy. ConMoE uses unlabeled calibration inputs and keeps the router and expert weights frozen. With the pool fixed, functional assignment improves macro accuracy by 1.82 and 0.35 percentage points at 50\% and 75\% expert retention, respectively, and also improves pools constructed by other selectors. With task-specific calibration across three MoE architectures and six zero-shot benchmarks, ConMoE achieves the highest macro average among the compared methods at both ratios. At 50\% expert retention, ConMoE reduces peak GPU memory by 29.6--48.4\% relative to the uncompressed models on our generation workloads. These results establish expert assignment as an independent optimization axis for MoE compression.
\end{abstract}

%% file: Body/1.intro.tex
\section{Introduction}
\label{sec:introduction}

Sparse Mixture-of-Experts (MoE) models scale model capacity while activating only a small subset of experts for each token~\citep{switch,olmoe,qwen3}. However, standard GPU-resident deployment still requires storing all model parameters in GPU memory. This has motivated growing interest in reducing the deployment memory footprint of MoE models while preserving their original capabilities.

Existing expert compression methods mainly fall into three categories: expert pruning, expert merging, and expert substitution. Expert pruning removes less important experts and restricts routing to the remaining ones~\citep{notall,reap}. Expert merging combines multiple experts into a smaller number of experts by aggregating their parameters~\citep{msmoe,hcsmoe}. Expert substitution approximates removed experts' outputs using lightweight substitutes, such as MoNE's novices~\citep{mone}. These approaches offer different ways to construct a compact pool of experts for deployment.

In this work, we formulate expert compression as two decisions: constructing an expert pool and assigning the original experts' routed workloads to that pool. Existing methods typically couple assignment with pool construction: pruning discards the outputs of removed experts, while merging maps each constituent expert to its merged representative. We observe that these decisions can be decoupled. Once a compact expert pool has been constructed, optimizing its assignment can further improve model quality without changing the expert pool or the compression budget.

\begin{figure}[t]
\centering
\includegraphics[width=\linewidth]{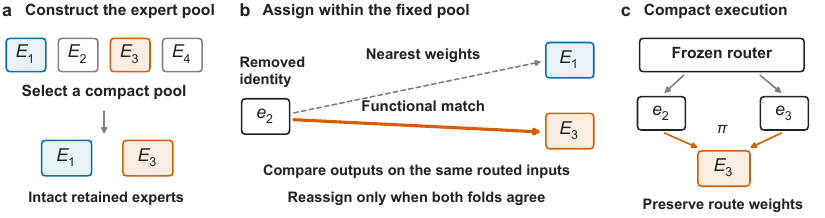}
\caption{\textbf{ConMoE separates pool construction from assignment.} (a) Select intact experts with Anchor Coverage. (b) Fix the pool and accept a replacement when both calibration folds select the same candidate and estimate a positive gain over the nearest assignment. (c) Store each retained expert once while preserving logical route weights. Assignments are schematic.}
\label{fig:overview}
\end{figure}

We introduce \method\ (Figure~\ref{fig:overview}) to realize this decoupling. A bound on MoE-layer reconstruction error yields a router-weighted functional assignment objective; parameter proximity is only a proxy for that objective. ConMoE first constructs an expert pool using \emph{Anchor Coverage}, then selects replacements by comparing expert outputs on the inputs routed to each removed identity. Weight proximity provides a small candidate set for functional evaluation and a fallback. The procedure uses only forward computation on unlabeled inputs and keeps the router and expert weights frozen.

Across three MoE architectures and six zero-shot benchmarks, changing only the assignment raises the macro average from 61.10\% to 62.91\% at 50\% retention, a gain of 1.82 percentage points; at 75\% retention, the gain is 0.35 points. The same assignment procedure improves Frequency and REAP pools by 4.73 and 1.75 points, respectively, at 50\% retention. These gains accompany better held-out expert and layer reconstruction. Under task-specific calibration, ConMoE achieves the highest macro average among the compared compressors at both 50\% and 75\% expert retention. At 50\% expert retention, ConMoE reduces peak GPU memory by 29.6--48.4\% relative to the uncompressed models on our generation workloads.

Our contributions are threefold: (i) we formulate expert compression as pool construction and assignment, exposing an opportunity to optimize assignment within a fixed pool; (ii) we develop ConMoE, combining Anchor Coverage with router-conditioned functional assignment derived from a reconstruction bound; and (iii) we establish the benefits through fixed-pool comparisons, held-out reconstruction, and physical compaction across three MoE architectures. Section~\ref{sec:related} places the two decisions in prior work; Sections~\ref{sec:formulation} and~\ref{sec:method} isolate assignment after the pool is fixed.

%% file: Body/5.related.tex
\section{Related Work}
\label{sec:related}

\paragraph{Pruning and compact expert pools.}
Expert pruning selects retained experts using sensitivity, search, or activation statistics~\citep{unified,notall}. REAP combines router weights with activation norms~\citep{reap}, while AIMER ranks experts without calibration inputs~\citep{aimer}. These methods select the retained pool and drop removed contributions. EEP separately searches effective routing and merged expert representations through evolutionary optimization~\citep{eep}. ConMoE fixes the original router and an intact expert pool, then optimizes a discrete post-routing identity-to-expert assignment.

\paragraph{Merging and expert representation.}
M-SMoE aligns and merges experts using routing information~\citep{msmoe}; HC-SMoE clusters expert outputs~\citep{hcsmoe}. Sub-MoE merges components within a shared subspace~\citep{submoe}, and REAM combines activation-based selection with merging~\citep{ream}. $D^2$-MoE uses a shared base and compressed deltas~\citep{d2moe}, and LorExperts uses local dominants with low-rank corrections~\citep{lorexperts}. In these cases the assignment of a removed or merged identity is inherited from the grouping or factorization that built the new representation. ConMoE instead keeps intact original experts and treats assignment as a separate map onto that fixed pool.

\paragraph{Substitution and frozen-weight compression.}
MoNE is the closest prior method~\citep{mone}. It selects experts and replaces the rest with calibration-mean novice approximators, so selection and replacement are designed together. ConMoE stays in the substitution family but does not fit a new expert: a removed identity is remapped to an existing retained expert, and that remapping is optimized after the pool is fixed. Router KD studies router distillation with frozen experts~\citep{routerkd}. We freeze both the router and the expert weights; the only free object after pool construction is the assignment.

%% file: Body/2.formulation.tex
\section{MoE Compression as Pool Construction and Assignment}
\label{sec:formulation}

\subsection{Expert Pool Construction and Assignment}
\label{sec:pool_assignment}

Consider an MoE layer $\ell$ with expert set $\mathcal E_\ell$. Given a hidden state $h$, the router selects a set of experts $\mathcal T_\ell(h)$ and assigns each selected expert a routing weight $g_{\ell,e}(h)$. Let $E_{\ell,e}(h)$ denote the output of expert $e$ on this input. The layer combines these outputs as
\begin{equation}
y_\ell(h)=\sum_{e\in\mathcal T_\ell(h)}g_{\ell,e}(h)E_{\ell,e}(h).
\label{eq:original_moe}
\end{equation}
After compression, the model stores a smaller expert pool $\mathcal C_\ell$. For an expert $p$ in this pool, let $P_{\ell,p}(h)$ denote its output. An assignment $\pi_\ell(e)$ specifies which pool member serves in place of each original expert $e$. To represent pruning, we also allow $\pi_\ell(e)=\bot$, with $P_{\ell,\bot}(h)=0$; this zero-output target requires no stored parameters. The compressed layer then computes
\begin{equation}
\widetilde y_\ell(h)=\sum_{e\in\bar{\mathcal T}_\ell(h)}\bar g_{\ell,e}(h)P_{\ell,\pi_\ell(e)}(h),
\label{eq:compressed_moe}
\end{equation}
where $\bar{\mathcal T}_\ell(h)$ and $\bar g_{\ell,e}(h)$ describe the active original identities and their weights under the compressed routing rule. This allows routing and weight normalization to change with compression. In this view, pruning assigns removed experts to zero output, merging assigns group members to their merged representative, and substitution assigns them to retained experts or lightweight replacement modules.

ConMoE retains $K_\ell$ original experts, so $\mathcal C_\ell\subseteq\mathcal E_\ell$ and $P_{\ell,p}=E_{\ell,p}$. It preserves the router: $\bar{\mathcal T}_\ell(h)=\mathcal T_\ell(h)$ and $\bar g=g$ on the same input. Its assignment maps into $\mathcal C_\ell$, with retained identities mapping to themselves. Once this pool is fixed, we optimize the assignments of removed identities without changing the available experts or their parameters.

\subsection{Router-Conditioned Functional Assignment}
\label{sec:functional_assignment}

We seek assignments that preserve the original layer output on the inputs where each replacement is used. Let $k$ be the maximum number of experts selected per input. Since retained identities contribute no error, the Cauchy--Schwarz inequality gives
\begin{equation}
\|y_\ell(h)-\widetilde y_\ell(h)\|_2^2
\leq k\sum_{e\in\mathcal T_\ell(h)\setminus\mathcal C_\ell}g_{\ell,e}(h)^2
\|E_{\ell,e}(h)-E_{\ell,\pi_\ell(e)}(h)\|_2^2.
\label{eq:layer_bound}
\end{equation}
This local reconstruction bound motivates comparing expert outputs on routed inputs, weighted by their contribution to the layer. Let $\mathcal D_{\ell,e}$ be the input distribution conditioned on the router selecting expert $e$. With the pool and input distribution fixed, the expected bound separates across identities because each retained expert can serve multiple assignments. For a removed identity with positive routing probability, the resulting functional cost and its minimizing assignment are
\begin{align}
F_\ell(e,p)&=\mathbb E_{h\sim\mathcal D_{\ell,e}}\!\left[g_{\ell,e}(h)^2\|E_{\ell,e}(h)-E_{\ell,p}(h)\|_2^2\right],\label{eq:functional_cost}\\
\pi_\ell^{\mathrm{func}}(e)&\in\arg\min_{p\in\mathcal C_\ell}F_\ell(e,p).
\label{eq:functional_assignment}
\end{align}
Appendix~\ref{app:derivation} gives the expectation and separability derivation. Unlike weight proximity, $F_\ell(e,p)$ evaluates a substitute on the workload routed to $e$. This relation is directional because reversing the substitution changes the routed input distribution and weights. Parameter proximity is therefore useful for shortlisting and as a fallback, but it does not define the assignment. Section~\ref{sec:method} develops an efficient approximation to this objective.

%% file: Body/3.method.tex
\section{ConMoE: Pool Construction and Conservative Reassignment}
\label{sec:method}

ConMoE implements the two decisions in Section~\ref{sec:formulation} using unlabeled calibration inputs. Any fixed pool makes assignment well-defined; Anchor Coverage is our default constructor. We first construct an expert pool that balances expert importance and coverage. With this pool fixed, we compare candidate replacements on routed inputs and update an assignment when the improvement is supported by two calibration subsets.

\subsection{Anchor Coverage for Pool Construction}
\label{sec:pool_construction}

Under a budget of $K_\ell$ experts, we first assess how much each expert contributes to the original computation. Motivated by Equation~\ref{eq:layer_bound}, we define expert demand as accumulated routing-weighted output energy on calibration hidden states $\{h_t\}$:
\begin{equation}
w_{\ell,e}=\sum_{t:e\in\mathcal T_\ell(h_t)}g_{\ell,e}(h_t)^2\|E_{\ell,e}(h_t)\|_2^2.
\label{eq:expert_demand}
\end{equation}
This score reflects how often an expert is selected, its routing weights, and its output magnitude. To account for relationships among experts as well as their individual demand, we incorporate pairwise parameter distances into pool construction. For projection $m\in\mathcal M=\{\mathrm{gate},\mathrm{up},\mathrm{down}\},$ let $W_{\ell,e}^{(m)}$ be expert $e$'s weight matrix. We define
\begin{equation}
d_\ell(e,p)=\frac{1}{3}\sum_{m\in\mathcal M}
\frac{\|W_{\ell,e}^{(m)}-W_{\ell,p}^{(m)}\|_F}
{\tfrac12(\|W_{\ell,e}^{(m)}\|_F+\|W_{\ell,p}^{(m)}\|_F)+\epsilon},
\label{eq:expert_distance}
\end{equation}
where $\epsilon>0$ stabilizes the denominator. Normalization makes the projections comparable despite differences in their weight scale. We combine this distance with expert importance through the weighted coverage objective
\begin{equation}
\mathcal L_{\mathrm{pool}}(\mathcal C_\ell)=\sum_{e\in\mathcal E_\ell}w_{\ell,e}\min_{p\in\mathcal C_\ell}d_\ell(e,p),
\qquad |\mathcal C_\ell|=K_\ell.
\label{eq:pool_objective}
\end{equation}
The nearest retained expert determines each original expert's coverage cost, with larger contributions receiving greater weight. Equation~\ref{eq:pool_objective} thus favors experts that jointly cover high-demand regions of the original pool. Because this aggregate objective may nevertheless trade away an individually high-demand expert, \emph{Anchor Coverage} protects a demand-based anchor set. We sort experts by decreasing $w_{\ell,e}$ and take the shortest prefix whose cumulative demand reaches a fraction $\rho=0.8$ of the total. If the prefix contains more than $K_\ell$ experts, we retain its $K_\ell$ highest-demand members. Otherwise, we initialize $\mathcal C_\ell$ with this prefix and greedily add the expert that most decreases Equation~\ref{eq:pool_objective} until the budget is filled. Expert identifiers break ties.

This stage fixes the physical pool and all of its parameters. Subsequent functional evidence can modify only the assignment map.

\subsection{Conservative Functional Reassignment}
\label{sec:conservative_assignment}\label{sec:assignment}

Within this pool, we seek replacements that minimize the functional cost in Equation~\ref{eq:functional_cost}. For every removed expert $e\notin\mathcal C_\ell$, we first initialize its assignment with the nearest retained expert under the parameter distance:
\begin{equation}
\pi_{\ell,0}(e)=\arg\min_{p\in\mathcal C_\ell}d_\ell(e,p).
\label{eq:fallback_assignment}
\end{equation}
This fallback is inexpensive once the distance matrix is available. Evaluating every retained expert would require many forward passes, so we form a candidate set $\mathcal S_\ell(e)\subseteq\mathcal C_\ell$ containing the fallback and nearby experts in weight space, with at most four candidates. Our implementation approximates the neighborhood using normalized fingerprints of 2,048 sampled parameters per projection; Appendix~\ref{app:fingerprint} compares accuracy and construction cost with full-weight retrieval. Parameter proximity narrows the search; routed expert outputs determine which replacement is accepted.

To obtain inputs representative of compressed execution, we run a base compressed model using the fixed pool and fallback assignments. For each removed expert, we collect up to 64 hidden states on which the original router selects its identity. The original expert supplies reference outputs on these same states. A candidate $p$ is evaluated through the empirical form of Equation~\ref{eq:functional_cost}:
\begin{equation}
\begin{aligned}
\ell_i(e,p)&=g_{\ell,e}(h_i)^2\|E_{\ell,e}(h_i)-E_{\ell,p}(h_i)\|_2^2,\\
\Delta_i(e,p)&=\ell_i(e,\pi_{\ell,0}(e))-\ell_i(e,p).
\end{aligned}
\label{eq:paired_gain}
\end{equation}
Every candidate is compared with the fallback on the same routed state. Positive $\Delta_i(e,p)$ means lower reconstruction error, giving a common decision boundary of zero.

Candidate rankings estimated from few observations may be unstable. We therefore require two disjoint subsets to support the same replacement. We implement the functional assignment objective in Equation~\ref{eq:functional_assignment} with a conservative two-fold selection rule. Ordered observations are assigned alternately to folds $\mathcal I_{e,0}$ and $\mathcal I_{e,1}$. Each fold selects the alternative with the largest mean improvement:
\begin{equation}
\begin{aligned}
G_f(e,p)&=\frac{1}{|\mathcal I_{e,f}|}\sum_{i\in\mathcal I_{e,f}}\Delta_i(e,p),\\
p_f^\star&\in\arg\max_{p\in\mathcal S_\ell(e)\setminus\{\pi_{\ell,0}(e)\}}G_f(e,p),\qquad f\in\{0,1\}.
\end{aligned}
\label{eq:fold_gain}
\end{equation}
We accept the replacement only when both folds choose the same expert and both estimate a positive gain:
\begin{equation}
\pi_\ell(e)=\begin{cases}
p^\star,&p_0^\star=p_1^\star=p^\star,\quad G_0(e,p^\star)>0,\;G_1(e,p^\star)>0,\\
\pi_{\ell,0}(e),&\text{otherwise}.
\end{cases}
\label{eq:consensus}
\end{equation}
Empty folds, a missing alternative, disagreement, or nonpositive gain preserve $\pi_{\ell,0}(e)$. Retained experts continue to map to themselves, and expert identifiers break ties. The common decision threshold is zero, with no task-specific gain threshold or reassignment quota. Because the pool remains fixed, the experiments can isolate the contribution of reassignment.

\subsection{Compact Execution}
\label{sec:compact_execution}

Once the assignments are determined, the compressed model stores the unchanged non-expert parameters, one copy of each retained expert, and the integer maps $\{\pi_\ell\}$. Each router-selected identity invokes its assigned expert with its original coefficient. If $\pi_\ell(e_1)=\pi_\ell(e_2)=p$, their contributions sum to $(g_{\ell,e_1}(h)+g_{\ell,e_2}(h))E_{\ell,p}(h)$. The implementation preserves the original slot order when accumulating these contributions, matching the reference execution of the same compressed assignment.

The weight-distance matrix is reusable across pools and tasks. Demand collection and functional evaluation require only forward computation, and all assignment decisions are completed before deployment. The router, retained experts, and other model parameters remain unchanged.

%% file: Body/4.experiments.tex
\section{Experiments}
\label{sec:experiments}

We first compare complete compressors across three MoE
architectures, then use controlled interventions to separate the contribution
of assignment from the choice of retained experts.

\subsection{Experimental Setup}
\label{sec:experimental_setup}

\paragraph{Models and evaluation.}
We evaluate Qwen3-30B-A3B~\citep{qwen3}, DeepSeek-MoE-16B~\citep{deepseek}, and OLMoE-1B-7B-0125~\citep{olmoe} on six zero-shot benchmarks: WinoGrande~\citep{winogrande}, ARC-Challenge and ARC-Easy~\citep{arc}, HellaSwag~\citep{hellaswag}, MMLU~\citep{mmlu}, and PIQA~\citep{piqa}. Evaluation uses the full splits with the Language Model Evaluation Harness. We report normalized accuracy for ARC-C, ARC-E, HellaSwag, and PIQA, and accuracy for MMLU and WinoGrande. Model averages weight the six tasks equally, and macro averages weight the three models equally.

\paragraph{Compression protocol.}
We focus on 50\% routed-expert reduction (50\% retention), with results at 25\% reduction (75\% retention) reported in Appendix~\ref{app:reduction25}. For the main comparison and fixed-pool experiments, we calibrate one compressed model per task using 128 unlabeled prompts, each capped at 2,048 tokens and disjoint from evaluation. Section~\ref{sec:practical_properties} evaluates shared calibration for a single compressed model serving all tasks. For ConMoE, the router and expert weights remain frozen, with no gradients or fine-tuning. Appendix~\ref{app:implementation} specifies calibration sampling and the common four-candidate, 64-state configuration.

\paragraph{Baselines.}
We compare three families of MoE compression methods:
expert pruning (frequency-based pruning and REAP),
expert merging (M-SMoE and HC-SMoE), and expert substitution
(MoNE). The uncompressed checkpoint serves as a reference.
Comparisons use the same pretrained backbone at each
routed-expert reduction level. Frequency and REAP prune removed experts at both reduction ratios. Baseline implementations and architecture adaptations are detailed in Appendix~\ref{app:baselines}.
Reduction ratios refer to the physical routed-expert pool;
auxiliary parameters are included when reporting total
model sizes.

\subsection{Main Results}
\label{sec:main_results}

\paragraph{Overall performance.}
Table~\ref{tab:main-results-t6} presents the six-task results at 50\% expert reduction. ConMoE attains the highest three-model macro average among the compared methods, 62.91\%, slightly above MoNE. At 25\% reduction, ConMoE also achieves the highest macro average, 67.31\% (Appendix~\ref{app:reduction25}).

\paragraph{Architecture dependence.}
ConMoE attains the highest compressed model average on Qwen3 and DeepSeek at 50\% expert reduction. MoNE leads on OLMoE, scoring 58.90 versus 58.03. We next separate the contributions of pool construction and assignment through controlled comparisons.

\input{Body/main_table}

\subsection{Disentangling Pool Construction and Assignment}
\label{sec:assignment_ablation}

\paragraph{Pool construction.}
To isolate pool construction, we compare Frequency, REAP, and Anchor Coverage as pool selectors, assigning each removed identity to its parameter-nearest retained expert. Anchor Coverage achieves a macro average of 61.10, compared with 54.06 for Frequency and 58.13 for REAP, and leads on all three models (Table~\ref{tab:assignment_ablation}). Pool construction therefore improves quality before functional reassignment. Appendix~\ref{app:anchor-threshold} examines the choice of anchor threshold.

\begin{figure}[t]
\centering
\includegraphics[width=\linewidth]{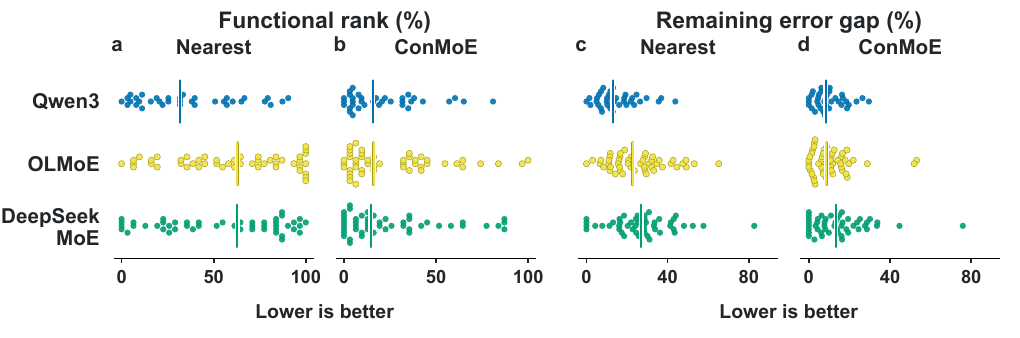}
\caption{\textbf{Functional assignment improves reconstruction on held-out inputs.} Each point is a removed expert identity at 50\% retention, evaluated on ARC-C or MMLU; vertical lines mark medians (Qwen3: 35 identities; OLMoE and DeepSeek: 48 each). (a,b) Functional rank among all retained candidates in the same layer, normalized from 0 (best) to 100 (worst). (c,d) Remaining MSE above the best candidate on the evaluation inputs, divided by weight-nearest MSE for both methods. Lower is better throughout. ConMoE's assignments are frozen before evaluation on disjoint inputs; the retained pool is fixed.}
\label{fig:functional-headroom}
\end{figure}

\paragraph{Held-out expert reconstruction.}
On ARC-C and MMLU at 50\% retention, we select ConMoE's assignments using 64 calibration prompts and freeze them before comparison with weight-nearest on 64 disjoint prompts from the same task. All retained candidates are evaluated on identical routed inputs for each sampled removed identity (Appendix~\ref{app:heldout}). ConMoE improves median functional ranks and error gaps to the best retained candidate across all three architectures (Figure~\ref{fig:functional-headroom}). Mean relative reconstruction-error reductions are positive in all six model--task settings, averaging 10.71\%.

\begin{figure}[t]
\centering
\includegraphics[width=\linewidth]{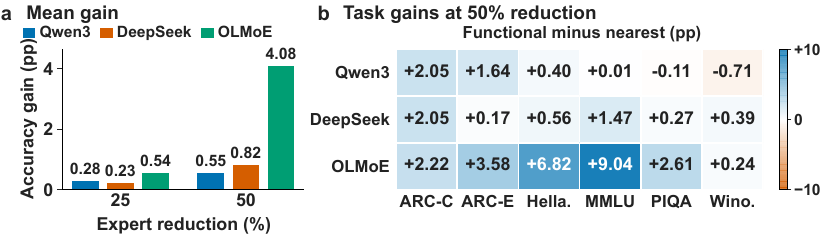}
\caption{\textbf{Fixed-pool assignment gains.} (a) Six-task mean accuracy gains at the two compression levels. (b) Task-level gains at 50\% expert reduction. Gains compare functional with nearest assignment within fixed Anchor Coverage pools, in percentage points.}
\label{fig:assignment-detail}
\end{figure}

\paragraph{Fixed-pool reassignment.}
To assess downstream benefits, we change only the identity-to-expert mapping, holding the expert pool, weights, router, and calibration data fixed (Table~\ref{tab:assignment_ablation}). Functional reassignment improves Anchor Coverage's macro accuracy by 1.82 percentage points at 50\% reduction, with gains on all three models and 16 of 18 model--task pairs. The gain is 0.35 points at 25\% reduction (Appendix~\ref{app:reduction25}); Figure~\ref{fig:assignment-detail} shows how gains vary with compression and task. Frequency and REAP pools also benefit, with macro gains of 4.73 and 1.75 points. These gains across selectors show that compression quality depends on how routed workloads use the retained pool. Appendix~\ref{app:uncertainty} reports an independent paired evaluation with bootstrap confidence intervals.

\input{Body/ablation_table}

\subsection{Layer Reconstruction and Prediction Fidelity}
\label{sec:functional_analysis}

We next examine layer reconstruction and prediction fidelity using the same 64/64 split on ARC-C and MMLU across all three architectures. Errors are measured against the uncompressed model, using common inputs for layer outputs and end-to-end execution for predictions (Appendix~\ref{app:layer-reconstruction}). Table~\ref{tab:functional_analysis}(a) shows that layer-output error decreases in all six settings, by 17.19\% on average. Prediction KL decreases in five of six settings, with a mean reduction of 12.34\%.

These improvements accompany higher downstream accuracy in five of the six settings (Table~\ref{tab:functional_analysis}(b)), with a mean gain of 2.64 percentage points and a 9.14-point gain on OLMoE MMLU. Functional reassignment therefore preserves computation at multiple levels while improving task performance. Appendix~\ref{app:controls} provides the full diagnostics and comparisons with alternative assignment rules.

\input{Body/functional_table}

\subsection{Practical Properties}
\label{sec:practical_properties}

\paragraph{Calibration sensitivity.}
To evaluate a single compressed model serving multiple tasks, we calibrate each architecture on WikiText or C4 and compare assignments within the resulting fixed pool. Table~\ref{tab:calibration_sources} shows positive gains for every model, with macro improvements of 1.21 and 2.00 percentage points, respectively. Functional reassignment thus remains effective with shared calibration data. Gains also persist across three calibration samples; Appendix~\ref{app:calibration} reports these results and further calibration and configuration analyses.

\input{Body/calibration_table}

\paragraph{Shared WikiText calibration.}
Shared-source calibration yields lower absolute accuracy than task-specific calibration (Table~\ref{tab:calibration_sources}). Because MoE routes different inputs to different experts, generic calibration may not reflect the expert usage of each downstream task. Prior work likewise finds that domain-specific calibration improves compressed-model quality~\citep{reap}. Table~\ref{tab:generic-complete} details the six-task comparison on Qwen3 under shared WikiText-128 calibration at 50\% retention. ConMoE achieves the highest average, 66.80\%, compared with 65.47\% for MoNE and 64.45\% for REAP, and ranks first or second on every task.

\input{Body/generic_complete}

\paragraph{Physical compression.}
At 50\% expert reduction, physical compaction lowers peak device memory by 29.6--48.4\% and achieves 1.11--1.24$\times$ the original model's throughput across batches 1, 8, and 32, measured over three repetitions (Table~\ref{tab:deployment}). The resulting models contain 45.6--47.5\% fewer unique parameters. Appendix~\ref{app:runtime} specifies the generation workload and measurement procedure.

\input{Body/runtime_table}

%% file: Body/main_table.tex
\begin{table*}[t]
\centering
\small
\setlength{\tabcolsep}{3.0pt}
\renewcommand{\arraystretch}{1.02}
\caption{Zero-shot accuracy (\%) on six benchmarks at 50\% expert reduction; Avg.\ weights tasks equally. Reduction is the fraction of routed experts removed. Bold and underlined values mark the best and second-best compressed results within each model--reduction block.}
\label{tab:main-results-t6}

\begin{tabular}{@{}lllccccccc@{}}
\toprule
Model & Reduction & Method
& Wino. & ARC-C & ARC-E
& Hella. & MMLU & PIQA & Avg. \\
\midrule

\multirow{7}{*}{\makecell[l]{Qwen3-\\30B-A3B}}
& -- & Original
& 70.17 & 56.31 & 79.21
& 77.62 & 77.82 & 80.36 & 73.58 \\
\cmidrule(lr){2-10}

& \multirow{6}{*}{50\%}
& M-SMoE
& 68.27 & 51.02 & 75.88
& 72.33 & 62.59 & \textbf{80.20} & 68.38 \\
& & HC-SMoE
& 52.17 & 30.72 & 50.34
& 37.04 & 41.89 & 63.00 & 45.86 \\
& & Frequency
& 69.06 & 52.05 & \underline{77.40}
& 74.78 & 69.07 & 79.49 & 70.31 \\
& & REAP
& 68.59 & \textbf{54.95} & 76.60
& 73.67 & 69.73 & 79.05 & 70.43 \\
& & MoNE
& \underline{69.30} & 51.88 & \textbf{77.86}
& \textbf{76.47} & \underline{69.91}
& \underline{80.14} & \underline{70.93} \\
\addlinespace[1.5pt]
& & \textbf{ConMoE}
& \textbf{69.38} & \underline{54.10} & 77.36
& \underline{75.97} & \textbf{70.76}
& 79.60 & \textbf{71.19} \\
\midrule

\multirow{7}{*}{\makecell[l]{DeepSeek-\\MoE-16B}}
& -- & Original
& 70.01 & 45.73 & 69.61
& 77.26 & 38.46 & 79.76 & 63.47 \\
\cmidrule(lr){2-10}

& \multirow{6}{*}{50\%}
& M-SMoE
& 55.96 & 35.07 & 57.79
& 52.87 & 25.41 & 74.86 & 50.33 \\
& & HC-SMoE
& 64.88 & 37.12 & 60.90
& 62.93 & 29.40 & 74.65 & 54.98 \\
& & Frequency
& 62.83 & 39.25 & 62.63
& 67.89 & 31.06 & 77.31 & 56.83 \\
& & REAP
& 65.43 & \underline{40.96} & \textbf{67.89}
& \underline{69.99} & 28.96 & 78.07 & \underline{58.55} \\
& & MoNE
& \textbf{67.09} & 40.27 & 66.62
& 66.10 & \textbf{31.11}
& \underline{78.62} & 58.30 \\
\addlinespace[1.5pt]
& & \textbf{ConMoE}
& \underline{66.77} & \textbf{42.41} & \underline{67.30}
& \textbf{70.03} & \underline{31.07}
& \textbf{79.54} & \textbf{59.52} \\
\midrule

\multirow{7}{*}{\makecell[l]{OLMoE-\\1B-7B-0125}}
& -- & Original
& 68.98 & 49.23 & 77.02
& 78.23 & 53.60 & 79.65 & 67.79 \\
\cmidrule(lr){2-10}

& \multirow{6}{*}{50\%}
& M-SMoE
& 53.43 & 31.06 & 55.35
& 33.13 & 33.46 & 63.11 & 44.92 \\
& & HC-SMoE
& 56.43 & 36.43 & 60.69
& 52.86 & 33.03 & 67.19 & 51.11 \\
& & Frequency
& 57.06 & 34.39 & 54.71
& 49.48 & 33.60 & 70.46 & 49.95 \\
& & REAP
& 57.54 & 34.30 & 58.29
& 56.37 & 30.44 & 71.00 & 51.32 \\
& & MoNE
& \textbf{62.83} & \textbf{42.15} & \underline{67.55}
& \underline{59.83} & \textbf{44.18}
& \textbf{76.88} & \textbf{58.90} \\
\addlinespace[1.5pt]
& & \textbf{ConMoE}
& \underline{59.91} & \underline{41.98} & \textbf{68.01}
& \textbf{61.97} & \underline{40.83}
& \underline{75.46} & \underline{58.03} \\
\bottomrule
\end{tabular}
\end{table*}

%% file: Body/ablation_table.tex
\begin{table*}[t]
\centering
\footnotesize
\setlength{\tabcolsep}{4.5pt}
\renewcommand{\arraystretch}{1.08}

\caption{Pool construction and assignment ablation at 50\% expert reduction. Each Nearest--Functional pair shares the retained pool. Scores are six-task means (\%); Macro Avg.\ weights models equally. $\Delta$ is the paired macro gain in percentage points, computed before rounding. Reduction refers to routed experts.}
\label{tab:assignment_ablation}\label{tab:ablation}

\begin{tabular}{@{}lclrrrrr@{}}
\toprule
Pool selector
& Reduction
& Assignment
& \makecell{Qwen3-\\30B-A3B}
& \makecell{DeepSeek-\\MoE-16B}
& \makecell{OLMoE-\\1B-7B-0125}
& \makecell{Macro\\Avg.}
& $\Delta$ \\
\midrule

\multirow{2}{*}{\makecell[l]{Anchor Coverage\\(ours)}}
& \multirow{2}{*}{50\%}
& Nearest
& 70.65 & 58.70 & 53.94 & 61.10 & -- \\
& & Functional
& 71.19 & 59.52 & 58.03 & 62.91 & +1.82 \\
\midrule

\multirow{2}{*}{Frequency}
& \multirow{2}{*}{50\%}
& Nearest
& 64.10 & 55.47 & 42.61 & 54.06 & -- \\
& & Functional
& 68.64 & 55.74 & 51.98 & 58.79 & +4.73 \\
\midrule

\multirow{2}{*}{REAP}
& \multirow{2}{*}{50\%}
& Nearest
& 68.60 & 57.08 & 48.72 & 58.13 & -- \\
& & Functional
& 69.02 & 57.23 & 53.41 & 59.89 & +1.75 \\
\bottomrule
\end{tabular}
\end{table*}

%% file: Body/functional_table.tex
\begin{table*}[!htbp]
\centering
\footnotesize
\renewcommand{\arraystretch}{1.15}

\caption{Fixed-pool reassignment at 50\% retention across three models on ARC-C and MMLU. (a) Mean relative error changes from nearest to functional assignment on held-out inputs, using uncompressed outputs as references; negative values are better. ``Improved'' counts settings with lower error. (b) Functional-minus-nearest accuracy gains on full downstream evaluation sets, in percentage points.}
\label{tab:functional_analysis}

\begin{minipage}[t]{0.50\textwidth}
\centering
\vspace{0pt}
\textbf{(a) Reconstruction and prediction}
\par\vspace{5pt}

\setlength{\tabcolsep}{3pt}
\begin{tabular*}{\linewidth}
{@{\extracolsep{\fill}}lrr@{}}
\toprule
Metric & Rel.\ change & Improved \\
\midrule
Layer-output error         & $-17.19\%$ & 6/6 \\
Prediction KL              & $-12.34\%$ & 5/6 \\
Centered-logit MSE         & $-12.23\%$ & 6/6 \\
Top-1 disagreement         & $-7.71\%$  & 5/6 \\
\midrule
Expert reconstruction MSE & $-17.43\%$ & 6/6 \\
\bottomrule
\end{tabular*}
\end{minipage}
\hfill
\begin{minipage}[t]{0.47\textwidth}
\centering
\vspace{0pt}
\textbf{(b) Downstream accuracy gains}
\par\vspace{5pt}

\setlength{\tabcolsep}{3pt}
\begin{tabular*}{\linewidth}
{@{\extracolsep{\fill}}lrr@{}}
\toprule
Model & ARC-C & MMLU \\
\midrule
Qwen3-30B-A3B
& $+1.024$ & $-0.043$ \\
DeepSeek-MoE-16B
& $+1.962$ & $+1.467$ \\
OLMoE-1B-7B-0125
& $+2.304$ & $+9.137$ \\
\midrule
Mean / improved
& \multicolumn{2}{c}{$+2.642$ ; 5/6} \\
\bottomrule
\end{tabular*}
\end{minipage}

\end{table*}

%% file: Body/calibration_table.tex
\begin{table*}[!htbp]
\centering
\footnotesize
\setlength{\tabcolsep}{5pt}
\renewcommand{\arraystretch}{1.08}

\caption{Calibration sensitivity at 50\% expert reduction. Each pair shares the pool. Scores are six-task means (\%); Macro Avg.\ weights models equally, and $\Delta$ is the paired macro gain (percentage points). Task-specific calibration uses a model per task; WikiText and C4 each use one shared model per architecture.}
\label{tab:calibration_sources}

\begin{tabular}{@{}llrrrrr@{}}
\toprule
Calibration & Assignment
& \makecell{Qwen3-\\30B-A3B}
& \makecell{DeepSeek-\\MoE-16B}
& \makecell{OLMoE-\\1B-7B-0125}
& \makecell{Macro\\Avg.}
& $\Delta$ \\
\midrule

\multirow{2}{*}{Task-specific}
& Nearest
& 70.65 & 58.70 & 53.94 & 61.10 & -- \\
& Functional
& 71.19 & 59.52 & 58.03 & 62.91 & +1.82 \\
\midrule

\multirow{2}{*}{WikiText}
& Nearest
& 66.54 & 49.96 & 40.18 & 52.22 & -- \\
& Functional
& 66.80 & 50.97 & 42.54 & 53.44 & +1.21 \\
\midrule

\multirow{2}{*}{C4}
& Nearest
& 58.56 & 53.38 & 44.38 & 52.11 & -- \\
& Functional
& 61.21 & 54.75 & 46.36 & 54.11 & +2.00 \\
\bottomrule
\end{tabular}
\end{table*}

%% file: Body/generic_complete.tex
\begin{table}[!htbp]
\centering
\footnotesize
\setlength{\tabcolsep}{5pt}
\renewcommand{\arraystretch}{1.08}

\caption{Task-level comparison on Qwen3-30B-A3B with shared WikiText-128 calibration at 50\% expert retention. Each method produces one compressed model for all six tasks; REAP uses physical expert pruning. Scores are accuracies (\%), and Avg.\ weights the six tasks equally. Bold and underlined values mark the best and second-best results within each column.}
\label{tab:generic-complete}

\begin{tabular}{@{}lrrrrrrr@{}}
\toprule
Method
& ARC-C & ARC-E & Hella. & MMLU & PIQA & Wino. & Avg. \\
\midrule
REAP
& \underline{52.99} & \textbf{75.29} & 62.78 & 57.06 & 72.14 & 66.46 & 64.45 \\
MoNE
& 47.87 & 69.57 & \textbf{71.14} & \textbf{58.49} & \textbf{75.95} & \underline{69.77} & \underline{65.47} \\
\textbf{ConMoE}
& \textbf{53.07} & \underline{75.25} & \underline{68.85} & \underline{57.61} & \underline{75.84} & \textbf{70.17} & \textbf{66.80} \\
\bottomrule
\end{tabular}
\end{table}

%% file: Body/runtime_table.tex
\begin{table*}[!htbp]
\centering
\footnotesize
\setlength{\tabcolsep}{3.5pt}
\renewcommand{\arraystretch}{1.10}

\caption{Physical compression at 50\% expert reduction. Unique total parameters are in billions. Throughput and peak-memory changes compare compact and original execution over three repetitions.}
\label{tab:deployment}

\begin{tabular}{@{}lrrrrrrr@{}}
\toprule
& \multicolumn{3}{c}{Unique parameters}
& \multicolumn{3}{c}{Throughput ratio}
& \\
\cmidrule(lr){2-4}
\cmidrule(lr){5-7}
Model
& Original & Compact & Reduction
& Batch 1 & Batch 8 & Batch 32
& \makecell{Peak-memory\\change} \\
\midrule

Qwen3-30B-A3B
& 30.532 & 16.037 & 47.48\%
& 1.115$\times$ & 1.169$\times$ & 1.181$\times$
& $-48.38\%$ to $-47.30\%$ \\

DeepSeek-MoE-16B
& 16.376 & 8.901 & 45.64\%
& 1.123$\times$ & 1.221$\times$ & 1.237$\times$
& $-39.68\%$ to $-29.57\%$ \\

OLMoE-1B-7B-0125
& 6.919 & 3.698 & 46.56\%
& 1.240$\times$ & 1.177$\times$ & 1.199$\times$
& $-47.75\%$ to $-44.00\%$ \\
\bottomrule
\end{tabular}
\end{table*}

%% file: Body/6.conclusion.tex
\section{Conclusion}
\label{sec:conclusion}
We formulate MoE compression as expert pool construction and workload assignment, and introduce ConMoE to optimize both through Anchor Coverage and router-conditioned functional assignment. Functional assignment improves macro accuracy for all three pool selectors at both retention ratios. Held-out evaluations further show improved layer reconstruction and prediction fidelity. Under task-specific calibration, ConMoE achieves the highest macro accuracy among the compared compressors at both evaluated retention ratios, while substantially reducing GPU memory. These results establish assignment as a distinct component of MoE compression: quality depends on both which experts remain and how their capacity is used.

%% file: Body/appendix.tex
\section{Experimental and Implementation Details}
\label{app:implementation}

\subsection{Models and Calibration}
We use Qwen3-30B-A3B, DeepSeek-MoE-16B (the original DeepSeek-MoE architecture), and OLMoE-1B-7B-0125. The retained budget is $K_\ell=\operatorname{round}(r|\mathcal E_\ell|)$ for $r\in\{0.50,0.75\}$, applied separately within each layer. Shared experts, attention, embeddings, normalization, and router parameters are preserved.

Default calibration uses 128 deterministically sampled unlabeled prompts per task, with at most 2,048 tokens per prompt. ARC-C, ARC-E, and MMLU use validation examples for calibration and test examples for evaluation. HellaSwag, PIQA, and WinoGrande use training examples for calibration and validation examples for evaluation. Only prompt text is used in compression. Zero-shot evaluation uses the Language Model Evaluation Harness, with batch size 64 for Qwen3 and 32 for DeepSeek and OLMoE. MMLU aggregates its subjects using the harness task metric. Model means weight the six benchmarks equally.

The common configuration uses an 80\% demand nucleus, at most four functional candidates, at most 64 routed states per removed identity, and even/odd folds. No configuration is tuned separately by model or task. Retained identities map to themselves. A missing alternative, an empty fold, disagreement between folds, or nonpositive gain preserves the nearest assignment. Candidate ties are resolved by expert identifier. Both folds use a zero gain threshold, with no task-specific threshold or reassignment quota.

\subsection{Parameter Distance and Pool Construction}
\label{app:distance}
Let $W^{(a)}_{\ell,e}$ denote projection $a\in\{\mathrm{gate},\mathrm{up},\mathrm{down}\}$ of expert $e$. The pool objective and fallback use
\begin{equation}
d_\ell(e,p)=\frac{1}{3}\sum_a
\frac{\|W^{(a)}_{\ell,e}-W^{(a)}_{\ell,p}\|_F}
{\tfrac12(\|W^{(a)}_{\ell,e}\|_F+\|W^{(a)}_{\ell,p}\|_F)+\epsilon},
\qquad \epsilon=10^{-8}.
\end{equation}
We rank identities by demand and initialize the pool with the shortest prefix whose cumulative demand reaches 80\% of the total. When this nucleus exceeds the budget, we retain its $K_\ell$ highest-demand members. Greedy coverage fills any remaining positions, with expert identifiers breaking ties in both stages. Pairwise distances depend only on the checkpoint and can be reused across tasks and retained ratios. Functional observations are collected with the nearest-assignment compressed model held fixed; removed experts from the original checkpoint supply target outputs on those same inputs. All selected assignments are applied after collection.

\subsection{Baseline Implementations}
\label{app:baselines}
Frequency ranks experts by observed routing usage; REAP ranks experts by the mean router-weighted output norm over the tokens routed to each expert~\citep{reap}. In Tables~\ref{tab:main-results-t6} and~\ref{tab:main-results-r25}, both selectors prune removed experts at 50\% and 25\% reduction, respectively. In the fixed-pool comparisons, REAP uses the expert pool selected by its pruning baseline. Each nearest--functional pair preserves the pool, expert weights, and original router. The shared WikiText complete comparison also uses physical REAP pruning.

M-SMoE~\citep{msmoe} and HC-SMoE~\citep{hcsmoe} are implemented on the same backbones as ConMoE. These are architecture adaptations of the merging methods. MoNE~\citep{mone} retains its released float32 scoring recurrence, layer-local strict-cutoff selection, and calibration-mean novice substitution. Statistics are collected with unpadded batch size one. OLMoE uses a supported architecture branch; Qwen3 and the original DeepSeek-MoE use adapters preserving these operations. The main comparison reports routed-expert reduction, while ConMoE's deployment measurements report exact unique parameter counts.

\subsection{Physical Compaction and Runtime Measurement}
\label{app:runtime}
ConMoE stores one copy per retained expert and maps the original logical expert identities to this bank. Multiple routed slots may reference the same expert. The compact implementation preserves the original slot order and coefficients when accumulating contributions. We compare its logits with the reference execution of the same compressed assignment to verify numerical equivalence.

Generation measurements use 64 ARC-C prompts, inputs capped at 256 tokens, 32 greedy generated tokens, batches of 1, 8, and 32, one warm-up batch, and three repetitions per condition. Models run in bfloat16 with four visible GPUs. Throughput divides generated tokens by synchronized wall-clock time. Peak memory is the largest peak allocated memory among participating devices. Qwen3's original and compact measurements each use a fresh process, preventing placement from one condition from affecting the other.

\section{Derivation of the Reconstruction Objective}
\label{app:derivation}
For ConMoE, the retained pool satisfies $\mathcal C_\ell\subseteq\mathcal E_\ell$, retained identities map to themselves, and the original routing coefficients are preserved on a fixed input. Let $\mathcal R_\ell(h)=\mathcal T_\ell(h)\setminus\mathcal C_\ell$ be the active removed identities. For a fixed layer input $h$, write
$v_e=g_{\ell,e}(h)[E_{\ell,e}(h)-E_{\ell,\pi_\ell(e)}(h)]$.
Retained identities contribute zero error. If $s=|\mathcal R_\ell(h)|>0$, vector Cauchy--Schwarz yields
\begin{align}
\|y_\ell(h)-\widetilde y_\ell(h)\|_2^2
&=\left\|\sum_{e\in\mathcal R_\ell(h)}v_e\right\|_2^2
\leq s\sum_{e\in\mathcal R_\ell(h)}\|v_e\|_2^2\\
&\leq k\sum_{e\in\mathcal T_\ell(h)\setminus\mathcal C_\ell}
g_{\ell,e}(h)^2\|E_{\ell,e}(h)-E_{\ell,\pi_\ell(e)}(h)\|_2^2.
\end{align}
For $s=0$, both sides are zero. Let $\mathcal D_\ell$ denote the fixed distribution of layer inputs and $q_{\ell,e}=\Pr_{h\sim\mathcal D_\ell}[e\in\mathcal T_\ell(h)]$ the routing frequency. For $q_{\ell,e}>0$, conditioning $\mathcal D_\ell$ on this selection event gives $\mathcal D_{\ell,e}$ in Equation~\ref{eq:functional_cost}. Taking expectations gives
\begin{equation}
\E\|y_\ell(h)-\widetilde y_\ell(h)\|_2^2
\leq k\sum_{\substack{e\in\mathcal E_\ell\setminus\mathcal C_\ell\\q_{\ell,e}>0}}q_{\ell,e}F_\ell(e,\pi_\ell(e)).
\end{equation}
With the pool and input distribution fixed, the right-hand side is separable in $\pi_\ell(e)$: a retained expert may serve any number of identities, and there is no capacity constraint coupling their assignments. Positive routing frequencies scale each identity's cost without changing its minimizer, yielding Equation~\ref{eq:functional_assignment}. Identities with zero routing probability do not contribute to the expectation. In the implementation, identities without routed calibration observations retain their fallback. This motivates the bounded candidate search in Section~\ref{sec:assignment}. The result bounds local layer reconstruction on shared inputs; downstream measurements evaluate the composed compressed network.

\section{Additional Experimental Results}
\label{app:transfer}

\subsection{Complete Compression and Fixed-Pool Results}
\label{app:reduction25}
Table~\ref{tab:main-results-r25} reports complete-compressor results at 25\% expert reduction (75\% retention). ConMoE achieves the highest three-model macro average, 67.31\%, slightly above MoNE, and the highest compressed model averages on Qwen3 and DeepSeek. MoNE leads on OLMoE, scoring 66.03 versus 65.81. Table~\ref{tab:assignment-ablation-r25} reports fixed-pool comparisons across selectors. With Anchor Coverage, functional reassignment raises the macro average from 66.96 to 67.31, a gain of 0.35 percentage points.
\input{Body/main_table_r25}
\input{Body/ablation_table_r25}

\paragraph{Task-level fixed-pool comparisons.}
Tables~\ref{tab:app-assignment50} and~\ref{tab:app-assignment75} give the complete Anchor Coverage comparisons. Functional assignment improves 16 of 18 task cells at 50\% retention and 12 of 18 at 75\%. Tables~\ref{tab:app-frequency} and~\ref{tab:app-reap} report Frequency and REAP pools at both ratios. At 50\% retention, all model means improve under both selectors, with macro gains of 4.73 and 1.75 percentage points, respectively. At 75\% retention, the macro gains are 0.52 and 0.32 points; gains are concentrated in OLMoE, while Qwen3 declines under both selectors and DeepSeek declines slightly under REAP.
\input{Body/assignment50}
\input{Body/assignment75}
\input{Body/frequency}
\input{Body/reap}

\subsection{Assignment Analysis and Controls}
\label{app:controls}
\subsubsection{Held-Out Expert Reconstruction}
\label{app:heldout}
Figure~\ref{fig:functional-headroom} compares ConMoE and weight-nearest on ARC-C and MMLU across all three models at 50\% retention. We sample up to eight removed identities at each of three layers near normalized depths 0.2, 0.5, and 0.8, and separate assignment selection from evaluation. For each model and task, the first 64 calibration prompts determine ConMoE's assignment and the remaining 64 prompts evaluate both the frozen assignment and weight-nearest. The retained pool is constructed using the original 128 prompts and held fixed. We collect at most 64 routed states per sampled identity from the weight-nearest compressed model and evaluate every retained candidate on these common states. Of 144 sampled identities, 131 have routed evaluation inputs; the other 13 are excluded from both methods' summaries. The best evaluation candidate defines the diagnostic reference and does not alter ConMoE's assignment.

Among the 73 identities whose assignment changes, 66 improve and seven worsen; the remaining 58 keep their original mapping. Model medians pool supported identities across the two tasks, while mean relative error reductions weight identities equally within each model--task setting and then average the six settings equally. Median normalized rank falls from 31.7 to 15.9 on Qwen3, 62.9 to 16.1 on OLMoE, and 62.9 to 14.5 on DeepSeek, where lower is better.

\subsubsection{Layer Reconstruction and Prediction Fidelity}
\label{app:layer-reconstruction}
We use ARC-C and MMLU on all three models at 50\% retention. The 128 calibration prompts are divided into 64 construction prompts and 64 held-out prompts. The retained pool and candidate set are shared across assignment controls. Expert reconstruction evaluates removed and substitute expert outputs on identical routed hidden states. Layer reconstruction compares original and compressed MoE outputs on common reference inputs, normalized by reference output energy. End-to-end teacher-forced evaluation measures reference-to-compressed prediction KL, centered logit MSE, and top-one disagreement on valid prompt tokens. Task accuracy is evaluated on the complete downstream split.

Table~\ref{tab:mechanism-details} reports relative changes per setting. Each summary percentage averages the six relative changes, giving models and tasks equal weight. Figure~\ref{fig:functional-headroom} averages relative improvements over sampled identities, whereas Table~\ref{tab:functional_analysis} first aggregates expert errors by routed-state count within each setting. This differs from taking a ratio of the pooled raw metric means in Table~\ref{tab:controls}, which weights settings by their baseline error magnitude.
\input{Body/mechanism_details}

\subsubsection{Assignment Rules}
Nearest uses the parameter fallback. Single-fold selects the weighted functional minimizer using one construction fold. Two-fold agreement uses Equation~\ref{eq:consensus}. All-observation argmin minimizes weighted error over both folds without requiring agreement. Unweighted argmin removes the squared routing coefficient from that objective. Random selects a retained expert uniformly from the same candidate set. All controls preserve retained experts, routers, and weights.

\begin{figure}[H]
\centering
\includegraphics[width=\linewidth]{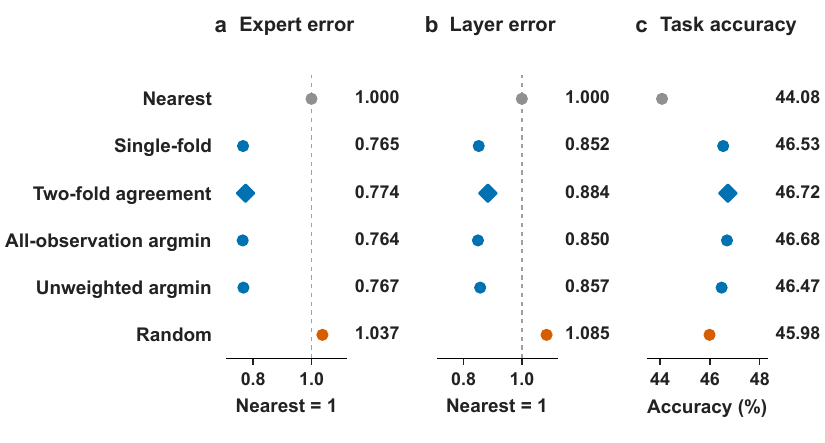}
\caption{\textbf{Assignment rules differ in reconstruction and downstream behavior.} All rules share the retained pool and candidate set on ARC-C and MMLU across three models at 50\% retention. (a,b) Mean expert and layer errors divided by their respective nearest-assignment means; the dashed baseline is one, and lower is better. This is a ratio of aggregate means, distinct from the mean per-setting reductions in Table~\ref{tab:functional_analysis}(a). (c) Mean task accuracy (\%), where higher is better. Diamonds mark two-fold agreement. Table~\ref{tab:controls} retains the raw metrics.}
\label{fig:assignment-controls}
\end{figure}

Figure~\ref{fig:assignment-controls} compares all six rules with the retained pool and candidates fixed. Every functional rule improves both expert and layer reconstruction. Random reassignment also improves mean task accuracy over nearest, but worsens both reconstruction metrics. Two-fold agreement has the highest downstream mean, 46.72\%, compared with 44.08\% for nearest and 45.98\% for random (Table~\ref{tab:controls}). All-observation argmin has the lowest expert and layer errors. These controls distinguish the preservation of routed computation from accuracy recovery through arbitrary reassignment, and support agreement as a competitive finite-sample decision rule.
\input{Body/controls}

\subsection{Calibration and Configuration Analysis}
\label{app:calibration}
\subsubsection{Calibration Samples}
Table~\ref{tab:seeds} compares the default deterministic sample with two additional samples using seeds 43 and 44. Pool construction and assignment are rerun for each sample, keeping the configuration fixed. Functional assignment improves all nine model--sample means at 50\% retention; Figure~\ref{fig:assignment-stability}(a) shows each sample separately.
\input{Body/seeds}

\subsubsection{Generic Calibration Sources}
WikiText-2~\citep{wikitext} and English C4~\citep{t5} each provide 128 training-source prompts with a 2,048-token cap. Each source produces one compressed model per architecture, shared across the six downstream tasks. Table~\ref{tab:calibration_sources} reports the fixed-pool assignment comparisons at 50\% expert retention. Table~\ref{tab:generic-complete} compares complete compressors on Qwen3 under the same WikiText-128 source; REAP uses physical pruning with a fixed 50\% retained-expert budget per layer.

\subsubsection{Token Budget and Longer Generic Calibration}
\input{Body/calibration_extra}

\subsubsection{Anchor Threshold}
\label{app:anchor-threshold}
We vary the cumulative demand threshold $\rho\in\{0,0.2,0.4,0.6,0.8,1.0\}$ across all three models and six tasks at 50\% expert retention. Calibration inputs, demand scores, parameter distances, and the nearest-assignment rule are held fixed. At $\rho=0$, coverage constructs the entire pool without anchors; its first expert minimizes total demand-weighted distance to the original experts. At $\rho=1$, selection reduces to the top-$K_\ell$ experts under the same demand score. We evaluate each distinct pool and mapping afresh and use the same result for thresholds yielding identical pools and mappings.

All sampled thresholds from 0 to 0.8 produce identical pools and assignments. Direct demand top-$K_\ell$ changes the pool in only three of the 18 model--task settings and yields nearly identical accuracy (Table~\ref{tab:anchor-threshold}). The default threshold and direct demand selection achieve macro averages of 61.040\% and 61.026\%, respectively. These results show limited sensitivity to the anchor threshold, with little additional accuracy benefit from coverage over demand-only selection in this setting.
\input{Body/anchor_threshold}

\subsubsection{Functional Evaluation Budget}
\label{app:sensitivity}
\input{Body/sensitivity}

\subsubsection{Paired Evaluation Uncertainty}
\label{app:uncertainty}
\begin{figure}[H]
\centering
\includegraphics[width=\linewidth]{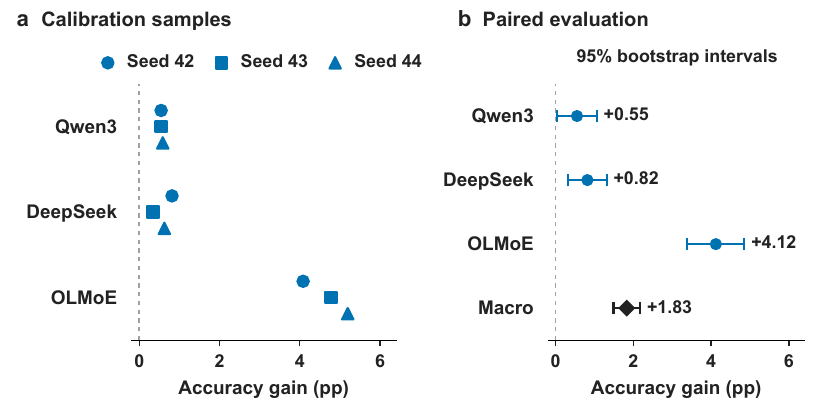}
\caption{\textbf{Assignment gains persist across calibration samples and paired evaluation.} All gains compare functional with nearest assignment at 50\% retention. (a) Six-task gains for three calibration samples per model, with pool construction and assignment rerun for each sample. (b) Independent paired evaluation: points are six-task model gains and the three-model macro gain; whiskers are percentile 95\% intervals from 10,000 paired bootstrap replicates. The panels report different sources of variation and use their respective evaluations. Tables~\ref{tab:seeds} and~\ref{tab:bootstrap} retain the numerical results.}
\label{fig:assignment-stability}
\end{figure}

We independently evaluate the nearest and functional assignments on matched examples at 50\% retention. For each model and task, we resample example indices with replacement, applying identical indices to both assignments. Each bootstrap replicate averages six task accuracies, then three model means for the macro result. We use 10,000 replicates and report percentile 95\% intervals. Table~\ref{tab:bootstrap} gives the paired evaluation results. Every model interval and the macro interval excludes zero. Figure~\ref{fig:assignment-stability} separates calibration-sample variation from these evaluation intervals: its left panel reruns compression with different samples, while its right panel resamples evaluation examples conditional on the compressed models.
\input{Body/bootstrap}

\subsection{Physical Compression and Deployment Efficiency}
\label{app:deployment-results}
Under the measurement procedure in Appendix~\ref{app:runtime}, compact execution has zero maximum logit difference from the reference execution of the same compressed assignment.

At 50\% retention, unique total parameters decrease from 30.532 to 16.037 billion for Qwen3, 16.376 to 8.901 billion for DeepSeek, and 6.919 to 3.698 billion for OLMoE. These correspond to reductions of 47.48\%, 45.64\%, and 46.56\%. Shared and non-expert parameters account for the difference between routed-expert retention and total parameter reduction.

At batches 1, 8, and 32, Qwen3's original throughput is 3.03, 7.46, and 17.97 tokens/s, versus 3.38, 8.73, and 21.22 for the compact model. Peak device memory decreases from 15,473, 15,552, and 15,828 MiB to 7,987, 8,083, and 8,341 MiB, respectively. Across the three batch sizes, peak memory reductions range from 29.6\% to 39.7\% for DeepSeek and 44.0\% to 47.8\% for OLMoE. These measurements describe the specified generation workload; the parameter reduction follows directly from the stored expert bank.

\section{Fingerprint Candidate Retrieval}
\label{app:fingerprint}
Our implementation uses sampled weight fingerprints to retrieve functional candidates. Controlled comparisons with full-parameter retrieval show higher average accuracy together with lower candidate construction time and temporary GPU memory.

\subsection{Implementation}
For each gate, up, and down projection, we take 2,048 evenly spaced entries from its flattened weights, normalize each sampled vector, and concatenate the three vectors. Thus each expert fingerprint contains 6,144 sampled parameters. Euclidean distance between these fingerprints ranks retained candidates. We insert the parameter-nearest fallback first, then add the closest distinct fingerprint candidates until four candidates have been selected or the retained pool is exhausted. Functional evaluation and the acceptance rule are unchanged.

\subsection{Accuracy Compared with Full-Parameter Retrieval}
Tables~\ref{tab:fp-accuracy} and~\ref{tab:fp-tasks} compare fingerprint retrieval with candidates retrieved using the full projection distance, holding the pool, fallback, candidate budget, routed samples, and assignment rule fixed. Fingerprints win 24 of 36 task comparisons and improve all six model--retention means, as visualized in Figure~\ref{fig:fingerprint-tradeoffs}(a). Across the 36 comparisons, mean accuracy rises from 64.8261\% to 65.1127\%, a gain of 0.2866 percentage points. The sampled implementation achieves higher average accuracy than full-parameter retrieval under the same assignment procedure.
\input{Body/fingerprint_accuracy}
\input{Body/fingerprint_tasks}

\subsection{Construction Time and Memory}
Table~\ref{tab:fp-cost} measures candidate-list construction after the model is loaded, with the ARC-C retained pool fixed at 50\%. Both implementations use vectorized, chunked operations with five alternating repetitions on NVIDIA A100-SXM4-80GB GPUs and eight CPU threads. Fingerprint construction uses chunks of 16 experts and 16 source identities. Timing excludes model loading, pool selection, functional replay, and serialization. Temporary GPU memory is allocation above the loaded-model baseline.

\begin{figure}[H]
\centering
\includegraphics[width=\linewidth]{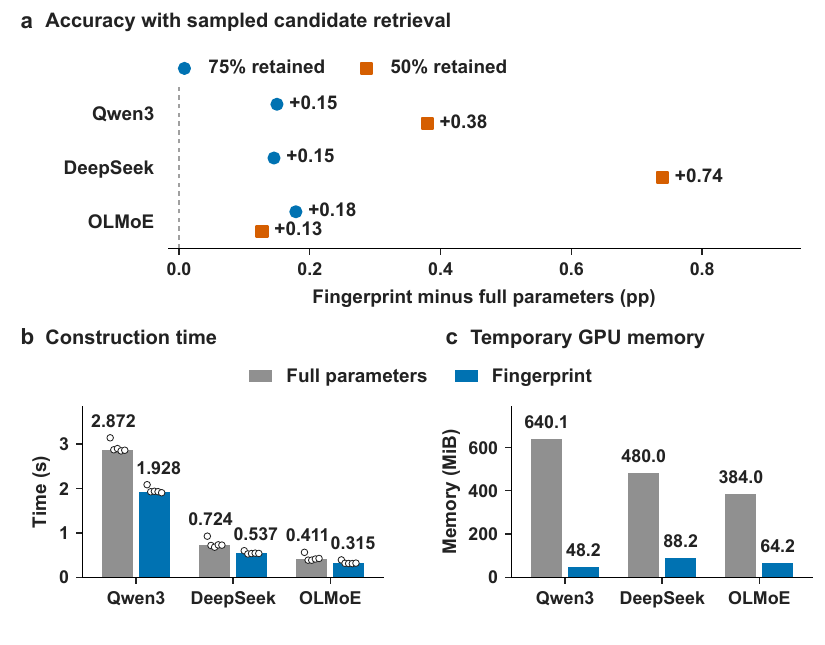}
\caption{\textbf{Sampled retrieval improves average accuracy and reduces candidate construction cost.} (a) Six-task accuracy gains over full-parameter retrieval under identical pools and assignment rules. (b) Construction time: medians, interquartile ranges, and individual measurements from five runs. (c) Median temporary GPU allocation above the loaded-model baseline. Cost measurements use the fixed ARC-C pool at 50\% retention. Tables~\ref{tab:fp-accuracy} and~\ref{tab:fp-cost} provide numerical results.}
\label{fig:fingerprint-tradeoffs}
\end{figure}

Figure~\ref{fig:fingerprint-tradeoffs}(b,c) shows that fingerprint construction is 1.31--1.49$\times$ faster and reduces temporary GPU memory by 81.6--92.5\%. The timing interquartile ranges, in seconds, are 2.862--2.899 versus 1.926--1.937 for Qwen3, 0.716--0.730 versus 0.534--0.543 for DeepSeek, and 0.385--0.424 versus 0.310--0.319 for OLMoE (full parameters versus fingerprints). These are preprocessing-stage savings for candidate construction. Deployment uses the completed integer map.
\input{Body/fingerprint_cost}

Completed candidate maps can be cached; over 100 loads, median loading times are approximately 1.84 ms for Qwen3, 0.52 ms for DeepSeek, and 0.29 ms for OLMoE for either retrieval method.

\subsection{Candidate Neighborhoods and Functional Substitutes}
To examine parameter and functional proximity, we sample up to eight removed identities at each of three layers near normalized depths 0.2, 0.5, and 0.8. We evaluate every retained substitute for each sampled identity on at most 64 routed states. The best substitute under this measured functional error defines a sampled functional oracle. The analysis covers ARC-C and MMLU on all three models at both retained ratios.

Table~\ref{tab:oracle} shows low overlap between the two candidate sets. Fingerprints have lower sampled-oracle recall but higher downstream mean accuracy (Table~\ref{tab:fp-accuracy}), so recovering the best measured substitute alone does not explain the accuracy gains. The different neighborhoods suggest candidate diversification as a possible contributor.
\input{Body/oracle}

\section{Limitations}
\label{app:limitations}
ConMoE requires unlabeled calibration data, and its compression quality depends on how well these data represent the target workloads. Although shared generic-text calibration supports a single compressed model across tasks, it yields lower accuracy than task-specific calibration in our experiments. This dependence is consistent with the conditional routing of MoE models: different input distributions activate different experts and expose different substitution behavior. A generic calibration corpus may therefore underrepresent expert workloads important to particular downstream tasks. Constructing calibration sets that better cover diverse workloads is an important direction for improving general-purpose compression.

%% file: Body/main_table_r25.tex
\begin{table*}[!htbp]
\centering
\small
\setlength{\tabcolsep}{3.0pt}
\renewcommand{\arraystretch}{1.02}
\caption{Zero-shot accuracy (\%) on six benchmarks at 25\% expert reduction; Avg.\ weights tasks equally. Reduction is the fraction of routed experts removed. Bold and underlined values mark the best and second-best compressed results within each model--reduction block.}
\label{tab:main-results-r25}

\begin{tabular}{@{}lllccccccc@{}}
\toprule
Model & Reduction & Method
& Wino. & ARC-C & ARC-E
& Hella. & MMLU & PIQA & Avg. \\
\midrule

\multirow{7}{*}{\makecell[l]{Qwen3-\\30B-A3B}}
& -- & Original
& 70.17 & 56.31 & 79.21
& 77.62 & 77.82 & 80.36 & 73.58 \\
\cmidrule(lr){2-10}

& \multirow{6}{*}{25\%}
& M-SMoE
& \textbf{71.03} & 55.29 & 78.87
& 77.51 & 76.13 & 80.41 & 73.21 \\
& & HC-SMoE
& 69.46 & 46.50 & 72.85
& 65.36 & 64.36 & 75.84 & 65.73 \\
& & Frequency
& \underline{70.56} & \underline{55.97} & 79.17
& \underline{77.66} & 76.89 & 80.47 & \underline{73.45} \\
& & REAP
& 69.85 & 55.03 & \underline{79.59}
& 76.85 & \underline{77.03} & 79.71 & 73.01 \\
& & MoNE
& 69.53 & 55.46 & 78.54
& \textbf{77.74} & 76.78 & \textbf{80.79} & 73.14 \\
\addlinespace[1.5pt]
& & \textbf{ConMoE}
& 70.17 & \textbf{56.14} & \textbf{79.67}
& 77.43 & \textbf{77.20} & \underline{80.69}
& \textbf{73.55} \\
\midrule

\multirow{7}{*}{\makecell[l]{DeepSeek-\\MoE-16B}}
& -- & Original
& 70.01 & 45.73 & 69.61
& 77.26 & 38.46 & 79.76 & 63.47 \\
\cmidrule(lr){2-10}

& \multirow{6}{*}{25\%}
& M-SMoE
& 68.35 & 41.13 & 67.63
& 68.27 & 32.03 & 77.20 & 59.10 \\
& & HC-SMoE
& 69.14 & 41.81 & 66.79
& 74.01 & 34.28 & 78.51 & 60.76 \\
& & Frequency
& 69.14 & 43.26 & 68.10
& 74.77 & 34.96 & 79.11 & 61.55 \\
& & REAP
& 69.38 & \textbf{44.80} & \textbf{69.07}
& \textbf{76.59} & 35.02
& \underline{79.82} & \underline{62.44} \\
& & MoNE
& \underline{70.01} & 44.37 & \underline{68.90}
& 75.25 & \textbf{35.74} & 79.76 & 62.34 \\
\addlinespace[1.5pt]
& & \textbf{ConMoE}
& \textbf{70.09} & \underline{44.45} & \textbf{69.07}
& \underline{76.30} & \underline{35.66}
& \textbf{79.92} & \textbf{62.58} \\
\midrule

\multirow{7}{*}{\makecell[l]{OLMoE-\\1B-7B-0125}}
& -- & Original
& 68.98 & 49.23 & 77.02
& 78.23 & 53.60 & 79.65 & 67.79 \\
\cmidrule(lr){2-10}

& \multirow{6}{*}{25\%}
& M-SMoE
& 62.67 & 46.50 & 70.54
& 62.72 & 46.57 & 78.24 & 61.21 \\
& & HC-SMoE
& 66.22 & 45.14 & 72.98
& 71.23 & 45.73 & 76.28 & 62.93 \\
& & Frequency
& 66.54 & 45.22 & 70.20
& 66.87 & 48.63 & 77.53 & 62.50 \\
& & REAP
& \textbf{67.48} & 47.01 & 75.38
& 74.78 & \underline{49.10} & 77.91 & 65.28 \\
& & MoNE
& 66.61 & \textbf{48.89} & \textbf{76.18}
& \underline{75.35} & \textbf{50.35}
& \underline{78.78} & \textbf{66.03} \\
\addlinespace[1.5pt]
& & \textbf{ConMoE}
& \underline{67.40} & \underline{48.12} & \underline{75.88}
& \textbf{75.58} & 48.63
& \textbf{79.22} & \underline{65.81} \\
\bottomrule
\end{tabular}
\end{table*}

%% file: Body/ablation_table_r25.tex
\begin{table*}[!htbp]
\centering
\footnotesize
\setlength{\tabcolsep}{4.5pt}
\renewcommand{\arraystretch}{1.08}

\caption{Fixed-pool assignment ablation at 25\% expert reduction. Each Nearest--Functional pair shares the retained pool. Scores are six-task means (\%); Macro Avg.\ weights models equally. $\Delta$ is the paired macro gain in percentage points, computed before rounding. Reduction refers to routed experts.}
\label{tab:assignment-ablation-r25}

\begin{tabular}{@{}lclrrrrr@{}}
\toprule
Pool selector
& Reduction
& Assignment
& \makecell{Qwen3-\\30B-A3B}
& \makecell{DeepSeek-\\MoE-16B}
& \makecell{OLMoE-\\1B-7B-0125}
& \makecell{Macro\\Avg.}
& $\Delta$ \\
\midrule

\multirow{2}{*}{\makecell[l]{Anchor Coverage\\(ours)}}
& \multirow{2}{*}{25\%}
& Nearest
& 73.27 & 62.35 & 65.27 & 66.96 & -- \\
& & Functional
& 73.55 & 62.58 & 65.81 & 67.31 & +0.35 \\
\midrule
\multirow{2}{*}{Frequency}
& \multirow{2}{*}{25\%}
& Nearest
& 73.23 & 60.98 & 60.91 & 65.04 & -- \\
& & Functional
& 73.00 & 61.52 & 62.16 & 65.56 & +0.52 \\
\midrule
\multirow{2}{*}{REAP}
& \multirow{2}{*}{25\%}
& Nearest
& 72.80 & 61.93 & 64.68 & 66.47 & -- \\
& & Functional
& 72.43 & 61.85 & 66.09 & 66.79 & +0.32 \\
\bottomrule
\end{tabular}
\end{table*}

%% file: Body/assignment50.tex
\begin{table}[!htbp]
\centering
\caption{Anchor Coverage at 50\% retention: full fixed-pool assignment comparison. All scores are percentages.}
\label{tab:app-assignment50}
\small
\begin{tabular}{llrrrrrrr}
\toprule
Model & Assignment & ARC-C & ARC-E & Hella. & MMLU & PIQA & Wino. & Avg. \\
\midrule
Qwen3 & Nearest & 52.05 & 75.72 & 75.57 & 70.75 & 79.71 & 70.09 & 70.65 \\
 & Functional & 54.10 & 77.36 & 75.97 & 70.76 & 79.60 & 69.38 & 71.19 \\
DeepSeek & Nearest & 40.36 & 67.13 & 69.47 & 29.60 & 79.27 & 66.38 & 58.70 \\
 & Functional & 42.41 & 67.30 & 70.03 & 31.07 & 79.54 & 66.77 & 59.52 \\
OLMoE & Nearest & 39.76 & 64.44 & 55.15 & 31.79 & 72.85 & 59.67 & 53.94 \\
 & Functional & 41.98 & 68.01 & 61.97 & 40.83 & 75.46 & 59.91 & 58.03 \\
\bottomrule
\end{tabular}
\end{table}

%% file: Body/assignment75.tex
\begin{table}[!htbp]
\centering
\caption{Anchor Coverage at 75\% retention: full fixed-pool assignment comparison. All scores are percentages.}
\label{tab:app-assignment75}
\small
\begin{tabular}{llrrrrrrr}
\toprule
Model & Assignment & ARC-C & ARC-E & Hella. & MMLU & PIQA & Wino. & Avg. \\
\midrule
Qwen3 & Nearest & 56.23 & 79.12 & 77.24 & 76.88 & 80.52 & 69.61 & 73.27 \\
 & Functional & 56.14 & 79.67 & 77.43 & 77.20 & 80.69 & 70.17 & 73.55 \\
DeepSeek & Nearest & 44.54 & 69.49 & 76.07 & 34.98 & 79.43 & 69.61 & 62.35 \\
 & Functional & 44.45 & 69.07 & 76.30 & 35.66 & 79.92 & 70.09 & 62.58 \\
OLMoE & Nearest & 48.98 & 76.01 & 75.10 & 45.61 & 78.35 & 67.56 & 65.27 \\
 & Functional & 48.12 & 75.88 & 75.58 & 48.63 & 79.22 & 67.40 & 65.81 \\
\bottomrule
\end{tabular}
\end{table}

%% file: Body/frequency.tex
\begin{table}[!htbp]
\centering
\caption{Frequency fixed-pool assignment at 50\% and 75\% expert retention. Scores are accuracies (\%); Avg.\ weights the six tasks equally.}
\label{tab:app-frequency}
\small
\begin{tabular}{llrrrrrrr}
\toprule
Model & Assignment & ARC-C & ARC-E & Hella. & MMLU & PIQA & Wino. & Avg. \\
\midrule
\multicolumn{9}{l}{\textbf{50\% retention (50\% reduction)}} \\
Qwen3 & Nearest & 50.94 & 75.51 & 58.24 & 61.88 & 70.78 & 67.25 & 64.10 \\
 & Functional & 50.09 & 76.98 & 72.19 & 64.92 & 78.62 & 69.06 & 68.64 \\
DeepSeek & Nearest & 38.57 & 62.12 & 63.31 & 30.10 & 75.63 & 63.06 & 55.47 \\
 & Functional & 39.16 & 62.08 & 63.82 & 30.86 & 76.22 & 62.27 & 55.74 \\
OLMoE & Nearest & 34.39 & 40.70 & 35.71 & 30.57 & 61.81 & 52.49 & 42.61 \\
 & Functional & 39.59 & 60.86 & 45.97 & 37.17 & 71.55 & 56.75 & 51.98 \\
\midrule
\multicolumn{9}{l}{\textbf{75\% retention (25\% reduction)}} \\
Qwen3 & Nearest & 55.03 & 79.04 & 77.45 & 76.45 & 80.36 & 71.03 & 73.23 \\
 & Functional & 55.29 & 78.75 & 77.40 & 76.41 & 80.52 & 69.61 & 73.00 \\
DeepSeek & Nearest & 43.34 & 67.85 & 72.78 & 34.72 & 78.51 & 68.67 & 60.98 \\
 & Functional & 43.69 & 68.27 & 73.49 & 35.11 & 78.94 & 69.61 & 61.52 \\
OLMoE & Nearest & 47.53 & 71.55 & 59.27 & 44.77 & 76.61 & 65.75 & 60.91 \\
 & Functional & 46.93 & 71.63 & 62.33 & 47.28 & 78.40 & 66.38 & 62.16 \\
\bottomrule
\end{tabular}
\end{table}

%% file: Body/reap.tex
\begin{table}[!htbp]
\centering
\caption{REAP fixed-pool assignment at 50\% and 75\% expert retention. Scores are accuracies (\%); Avg.\ weights the six tasks equally.}
\label{tab:app-reap}
\small
\begin{tabular}{llrrrrrrr}
\toprule
Model & Assignment & ARC-C & ARC-E & Hella. & MMLU & PIQA & Wino. & Avg. \\
\midrule
\multicolumn{9}{l}{\textbf{50\% retention (50\% reduction)}} \\
Qwen3 & Nearest & 51.54 & 74.24 & 71.43 & 65.13 & 77.97 & 71.27 & 68.60 \\
 & Functional & 51.19 & 76.77 & 71.33 & 67.17 & 78.51 & 69.14 & 69.02 \\
DeepSeek & Nearest & 39.85 & 67.55 & 65.14 & 25.52 & 76.33 & 68.11 & 57.08 \\
 & Functional & 39.51 & 66.84 & 66.37 & 26.26 & 77.26 & 67.17 & 57.23 \\
OLMoE & Nearest & 32.68 & 58.04 & 49.91 & 25.10 & 69.48 & 57.14 & 48.72 \\
 & Functional & 37.12 & 61.36 & 59.24 & 29.23 & 72.80 & 60.69 & 53.41 \\
\midrule
\multicolumn{9}{l}{\textbf{75\% retention (25\% reduction)}} \\
Qwen3 & Nearest & 55.55 & 78.28 & 77.21 & 76.39 & 79.92 & 69.46 & 72.80 \\
 & Functional & 54.10 & 78.96 & 77.00 & 76.36 & 79.49 & 68.67 & 72.43 \\
DeepSeek & Nearest & 42.66 & 69.74 & 75.15 & 34.72 & 79.27 & 70.01 & 61.93 \\
 & Functional & 43.52 & 69.15 & 75.45 & 33.81 & 79.49 & 69.69 & 61.85 \\
OLMoE & Nearest & 48.72 & 75.72 & 74.09 & 44.33 & 77.64 & 67.56 & 64.68 \\
 & Functional & 49.57 & 77.78 & 75.30 & 47.56 & 77.58 & 68.75 & 66.09 \\
\bottomrule
\end{tabular}
\end{table}

%% file: Body/mechanism_details.tex
\begin{table}[!htbp]
\centering
\caption{Held-out error reductions (\%) and downstream accuracy gains (percentage points) for the fixed-pool mechanism experiment. Positive values indicate improvement.}
\label{tab:mechanism-details}
\small
\begin{tabular}{llrrrr}
\toprule
Model & Task & Expert MSE & Layer error & KL & Accuracy gain \\
\midrule
Qwen3 & ARC-C & +24.96 & +24.37 & -32.99 & +1.02 \\
Qwen3 & MMLU & +27.33 & +12.55 & +16.29 & -0.04 \\
OLMoE & ARC-C & +10.17 & +10.15 & +12.83 & +2.30 \\
OLMoE & MMLU & +9.51 & +9.72 & +38.32 & +9.14 \\
DeepSeek & ARC-C & +20.04 & +27.26 & +31.46 & +1.96 \\
DeepSeek & MMLU & +12.57 & +19.07 & +8.12 & +1.47 \\
\bottomrule
\end{tabular}
\end{table}

%% file: Body/controls.tex
\begin{table}[!htbp]
\centering
\caption{Assignment controls on ARC-C and MMLU across three models, at 50\% retention. Values average six settings; reconstruction, layer error, KL, and centered logit MSE are lower-is-better, and accuracy (\%) is higher-is-better.}
\label{tab:controls}
\small
\begin{tabular}{lrrrrr}
\toprule
Rule & Expert MSE & Layer error & KL & Logit MSE & Accuracy \\
\midrule
Nearest & 6.880 & 0.0586 & 0.5274 & 1.464 & 44.08 \\
Single-fold & 5.266 & 0.0500 & 0.3806 & 1.188 & 46.53 \\
Two-fold agreement & 5.325 & 0.0518 & 0.4115 & 1.241 & 46.72 \\
All-observation argmin & 5.258 & 0.0498 & 0.3832 & 1.191 & 46.68 \\
Unweighted argmin & 5.275 & 0.0503 & 0.3974 & 1.205 & 46.47 \\
Random & 7.138 & 0.0636 & 0.4574 & 1.417 & 45.98 \\
\bottomrule
\end{tabular}
\end{table}

%% file: Body/seeds.tex
\begin{table}[!htbp]
\centering
\caption{Six-task means under three calibration samples at 50\% retention. Each sample uses 128 unlabeled prompts per task.}
\label{tab:seeds}
\small
\begin{tabular}{llrrr}
\toprule
Sample seed & Model & Nearest & Functional & Gain (pp) \\
\midrule
42 & Qwen3 & 70.65 & 71.19 & +0.55 \\
42 & DeepSeek & 58.70 & 59.52 & +0.82 \\
42 & OLMoE & 53.94 & 58.03 & +4.08 \\
43 & Qwen3 & 70.46 & 71.00 & +0.55 \\
43 & DeepSeek & 58.79 & 59.14 & +0.35 \\
43 & OLMoE & 53.76 & 58.55 & +4.78 \\
44 & Qwen3 & 70.49 & 71.08 & +0.59 \\
44 & DeepSeek & 58.60 & 59.22 & +0.63 \\
44 & OLMoE & 53.41 & 58.61 & +5.20 \\
\bottomrule
\end{tabular}
\end{table}

%% file: Body/calibration_extra.tex
Matching prompt count does not equalize token count. The 128 C4 prompts contain 57,108 tokens for Qwen3, 58,964 for DeepSeek, and 58,418 for OLMoE, whereas task-specific calibration contains 4,276--25,315 tokens. We construct an equal-total control by shortening the same C4 documents to prefixes, preserving their order and matching the valid-token total separately for each model and task. Prefix lengths are allocated deterministically from document lengths alone. Table~\ref{tab:equal-tokens} reports the six-task results. Task-specific calibration outperforms this token-matched control in every model--task setting under both assignment rules.
\input{Body/equal_tokens}

We also use Zyda-2~\citep{zyda}, packing one common source slice into 100 blocks of 4,096 tokens with each model's tokenizer. This gives 409,600 calibration tokens per model and one compressed model per architecture and retained ratio. Table~\ref{tab:zyda} shows positive assignment gains for all six model--retention means.
\input{Body/zyda}

%% file: Body/equal_tokens.tex
\begin{table}[!htbp]
\centering
\caption{C4 calibration with the same valid-token total as each task-specific calibration set, at 50\% retention. Each model--task control is evaluated on its corresponding task. Values average six tasks.}
\label{tab:equal-tokens}
\small
\begin{tabular}{lrrr}
\toprule
Model & Nearest & Functional & Gain (pp) \\
\midrule
Qwen3 & 57.41 & 60.12 & +2.71 \\
DeepSeek & 51.93 & 53.67 & +1.74 \\
OLMoE & 44.50 & 46.46 & +1.96 \\
\bottomrule
\end{tabular}
\end{table}

%% file: Body/zyda.tex
\begin{table}[!htbp]
\centering
\caption{Assignment under Zyda-2 calibration: 100 blocks of 4,096 tokens per model. Scores are six-task mean percentages.}
\label{tab:zyda}
\small
\begin{tabular}{llrrr}
\toprule
Model & Retained & Nearest & Functional & Gain (pp) \\
\midrule
Qwen3 & 75\% & 71.86 & 72.56 & +0.70 \\
DeepSeek & 75\% & 61.23 & 61.94 & +0.71 \\
OLMoE & 75\% & 57.13 & 58.40 & +1.27 \\
Qwen3 & 50\% & 61.94 & 63.83 & +1.88 \\
DeepSeek & 50\% & 54.78 & 55.23 & +0.46 \\
OLMoE & 50\% & 39.45 & 42.20 & +2.75 \\
\bottomrule
\end{tabular}
\end{table}

%% file: Body/anchor_threshold.tex
\begin{table}[!htbp]
\centering
\caption{Anchor-threshold sensitivity at 50\% expert retention with nearest assignment. Model scores are six-task mean accuracies (\%); the macro average weights models equally. The endpoints use coverage without anchors ($\rho=0$) and direct demand top-$K_\ell$ selection ($\rho=1$).}
\label{tab:anchor-threshold}
\small
\begin{tabular}{lrrrr}
\toprule
Threshold $\rho$ & Qwen3 & DeepSeek & OLMoE & Macro \\
\midrule
0.0 & 70.611 & 58.701 & 53.808 & 61.040 \\
0.2 & 70.611 & 58.701 & 53.808 & 61.040 \\
0.4 & 70.611 & 58.701 & 53.808 & 61.040 \\
0.6 & 70.611 & 58.701 & 53.808 & 61.040 \\
0.8 (default) & 70.611 & 58.701 & 53.808 & 61.040 \\
1.0 & 70.611 & 58.638 & 53.830 & 61.026 \\
\bottomrule
\end{tabular}
\end{table}

%% file: Body/sensitivity.tex
To assess accuracy under different functional evaluation budgets on Qwen3, we vary candidate count, fingerprint samples per projection, and routed observations jointly around the common default (Table~\ref{tab:sensitivity}). All three configurations improve over nearest assignment. Increasing the budget does not monotonically improve accuracy: the six-task means are 71.36\%, 71.19\%, and 71.11\% for the small, default, and large settings, versus 70.65\% for their nearest control. The common default thus provides bounded functional evaluation without model- or task-specific configuration selection.
\input{Body/sensitivity_table}

%% file: Body/sensitivity_table.tex
\begin{table}[!htbp]
\centering
\caption{Qwen3 at 50\% retention under three common configurations, each used on all six tasks. The sampled entries column is per projection; total fingerprint dimension is three times this value.}
\label{tab:sensitivity}
\small
\begin{tabular}{lrrrr}
\toprule
Configuration & Candidates & Entries / projection & Routed states & Accuracy \\
\midrule
Nearest & -- & -- & -- & 70.65 \\
Small & 2 & 1,024 & 32 & 71.36 \\
Default & 4 & 2,048 & 64 & 71.19 \\
Large & 8 & 4,096 & 128 & 71.11 \\
\bottomrule
\end{tabular}
\end{table}

%% file: Body/bootstrap.tex
\begin{table}[!htbp]
\centering
\caption{Paired evaluation at 50\% retention with 10,000 paired bootstrap resamples. Accuracy is in percent; gains and confidence intervals are in percentage points.}
\label{tab:bootstrap}
\small
\begin{tabular}{lrrrr}
\toprule
Model & Nearest & Functional & Gain & 95\% interval \\
\midrule
Qwen3 & 70.64 & 71.19 & +0.55 & [+0.05, +1.06] \\
DeepSeek & 58.70 & 59.52 & +0.82 & [+0.31, +1.31] \\
OLMoE & 53.92 & 58.05 & +4.12 & [+3.39, +4.85] \\
Macro & 61.09 & 62.92 & +1.83 & [+1.49, +2.17] \\
\bottomrule
\end{tabular}
\end{table}

%% file: Body/fingerprint_accuracy.tex
\begin{table}[!htbp]
\centering
\caption{Fingerprint and full-parameter candidate retrieval under identical fixed pools and functional assignment. Accuracy is the six-task mean (\%); wins count tasks favoring fingerprints.}
\label{tab:fp-accuracy}
\small
\begin{tabular}{llrrrr}
\toprule
Model & Retained & Full parameters & Fingerprint & Gain (pp) & Wins / 6 \\
\midrule
Qwen3 & 75\% & 73.3998 & 73.5497 & +0.1499 & 4 \\
DeepSeek & 75\% & 62.4370 & 62.5824 & +0.1453 & 4 \\
OLMoE & 75\% & 65.6282 & 65.8070 & +0.1788 & 4 \\
Qwen3 & 50\% & 70.8133 & 71.1927 & +0.3794 & 3 \\
DeepSeek & 50\% & 58.7800 & 59.5193 & +0.7393 & 5 \\
OLMoE & 50\% & 57.8981 & 58.0251 & +0.1270 & 4 \\
\bottomrule
\end{tabular}
\end{table}

%% file: Body/fingerprint_tasks.tex
\begin{table}[!htbp]
\centering
\caption{Fingerprint minus full-parameter retrieval accuracy, in percentage points, for all 36 model--retention--task comparisons.}
\label{tab:fp-tasks}
\small
\begin{tabular}{llrrrrrr}
\toprule
Model & Retained & ARC-C & ARC-E & Hella. & MMLU & PIQA & Wino. \\
\midrule
Qwen3 & 75\% & +0.939 & -0.168 & +0.020 & +0.256 & +0.326 & -0.474 \\
DeepSeek & 75\% & +0.597 & -0.884 & +0.060 & -0.071 & +0.381 & +0.789 \\
OLMoE & 75\% & +0.171 & -1.305 & -0.408 & +0.669 & +0.762 & +1.184 \\
Qwen3 & 50\% & +0.597 & +1.768 & -0.050 & +0.228 & -0.109 & -0.158 \\
DeepSeek & 50\% & +1.451 & -0.463 & +0.757 & +0.520 & +0.435 & +1.736 \\
OLMoE & 50\% & +1.621 & -0.126 & +0.777 & +0.719 & +0.218 & -2.447 \\
\bottomrule
\end{tabular}
\end{table}

%% file: Body/fingerprint_cost.tex
\begin{table}[!htbp]
\centering
\caption{First candidate-list construction with the model already loaded and the ARC-C retained pool fixed at 50\%. Times are medians of five runs; memory is temporary device allocation above the loaded-model baseline.}
\label{tab:fp-cost}
\small
\begin{tabular}{lrrrrrr}
\toprule
Model & Full (s) & FP (s) & Speedup & Full (MiB) & FP (MiB) & Reduction \\
\midrule
Qwen3 & 2.872 & 1.928 & 1.49$\times$ & 640.13 & 48.20 & 92.5\% \\
DeepSeek & 0.724 & 0.537 & 1.35$\times$ & 480.03 & 88.20 & 81.6\% \\
OLMoE & 0.411 & 0.315 & 1.31$\times$ & 384.03 & 64.20 & 83.3\% \\
\bottomrule
\end{tabular}
\end{table}

%% file: Body/oracle.tex
\begin{table}[!htbp]
\centering
\caption{Sampled routed-functional oracle diagnostics. Percentages and Jaccard similarity are averaged equally over six model--task settings per retained ratio. Oracle recall is the fraction of sampled identities whose best retained functional substitute occurs in the candidate set.}
\label{tab:oracle}
\small
\begin{tabular}{lrrrr}
\toprule
Retained & Nearest mismatch (\%) & Full recall (\%) & FP recall (\%) & Jaccard \\
\midrule
75\% & 98.6 & 10.4 & 6.9 & 0.172 \\
50\% & 94.4 & 15.3 & 11.8 & 0.191 \\
\bottomrule
\end{tabular}
\end{table}